\pdfoutput=1

\documentclass[11pt]{article}

\usepackage{naacl2021}

\usepackage{times}
\usepackage{latexsym}
\usepackage{tcolorbox}

\usepackage[T1]{fontenc}

\usepackage[utf8]{inputenc}
\usepackage{graphicx}
\usepackage{amsmath}
\usepackage{multirow}
\usepackage{subfigure} 
\usepackage{array}

\usepackage{microtype}
\usepackage{enumitem}
\usepackage{color, soul}

\usepackage{booktabs}       

\newcommand{\colorback}[1]{\colorbox[HTML]{DAE8FC}{#1}}

\title{Noisy-Labeled NER with Confidence Estimation}

\author{Kun Liu$^1$\thanks{$\quad$Equal Contribution.} $\quad$ Yao Fu$^{2}$$^*$ $\quad$ Chuanqi Tan$^1$\thanks{$\quad$Corresponding author.} $\quad$ Mosha Chen$^1$ \\ 
  \textbf{Ningyu Zhang$^3$ $\quad$ Songfang Huang$^1$ $\quad$ Sheng Gao$^4$} \\
    $^1$Alibaba Group $\quad$ $^2$University of Edinburgh $\quad$ 
    $^3$Zhejiang University \\
    $^4$ Guizhou Provincial Key Laboratory of Public Big Data, Guizhou University \\
    \texttt{\{kun.liu624; sheng.gao.81\}@gmail.com} $\quad$ \texttt{yao.fu@ed.ac.uk}  \\
    \texttt{\{chuanqi.tcq; chenmosha.cms; songfang.hsf\}@alibaba-inc.com} \\
    \texttt{zhangningyu@zju.edu.cn}
}

\begin{document}
\maketitle
\begin{abstract}
Recent studies in deep learning have shown significant progress in named entity recognition (NER). 
Most existing works assume clean data annotation, yet a fundamental challenge in real-world scenarios is the large amount of noise from a variety of sources (e.g., pseudo, weak, or distant annotations). 
This work studies NER under a noisy labeled setting with calibrated confidence estimation. 
Based on empirical observations of different training dynamics of noisy and clean labels, we propose strategies for estimating confidence scores based on local and global independence assumptions.
We partially marginalize out labels of low confidence with a CRF model.
We further propose a calibration method for confidence scores based on the structure of entity labels. 
We integrate our approach into a self-training framework for boosting performance. 
Experiments in general noisy settings with four languages and distantly labeled settings demonstrate the effectiveness of our method
\footnote{Our code can be found at \url{https://github.com/liukun95/Noisy-NER-Confidence-Estimation}}. 
\end{abstract}

\section{Introduction}
\label{sec:intro}

\begin{figure}
		\centering
		\includegraphics[width=3in]{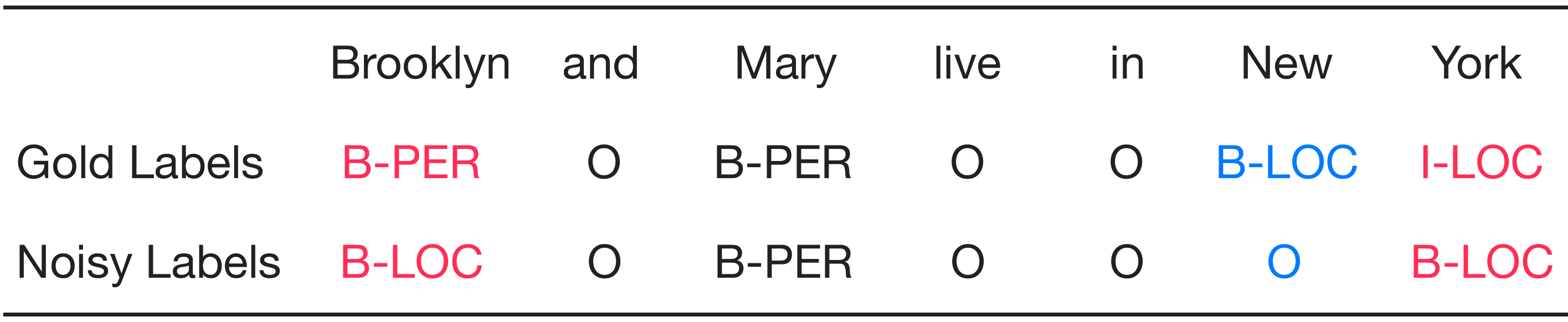}
		\caption{
		A noisy label example. \textit{Brooklyn} and \textit{York} are noisy positives.
		\textit{New} is  noisy negative.}
		\label{fig:example}
	\end{figure}

Recent progress in deep learning has significantly advanced NER performances~\citep{Lample2016, Devlin2018}. 
While most existing works assume clean data annotation,  real-world data inevitably involve different levels of noise (e.g., distant supervision from the dictionary~\citep{peng2019distantly}, or weak supervision from the web~\citealp{vrandevcic2014wikidata, cao-etal-2019-low}). 
Figure~\ref{fig:example} gives an example of such noisy labels. 
To train robust models with high performance, it is fundamentally critical to tackle the challenges associated with noisy data annotation.

In this work, we propose a confidence estimation approach for NER with noisy labels. 
We motivate our approach with important empirical observations of the training dynamics of clean and noisy labels: usually, clean data are easier to fit with faster convergence and smaller loss values~\citep{jiang2018mentornet,han2018coteaching,arazo2019unsupervised}.
Consequently, loss values (probabilities or scores of labels) can serve as strong indicators 
for the existence of noise, which we utilize to build our confidence estimation.

The key contribution of this work is a confidence estimation method with calibration. 
We use probabilities of labels as confidence scores and apply two estimation strategies based on global or local normalization that assume different  dependency structures about how the noisy labels are generated. 
We further calibrate the confidence score for positive labels (labels representing entity parts, e.g., \textit{B-LOC}) based on the structure of these labels: we separately estimate scores for the \textit{position part} (e.g., \textit{B} in \textit{B-LOC})  and the \textit{type part} (e.g., \textit{LOC} in \textit{B-LOC}). 
Such fine-grained calibration leads to a more accurate estimation and better performance in our experiments. 

We apply our method in a CRF model~\citep{Bellare2007, Yang2018}, marginalize out labels we do not trust, and maximize the likelihood of trusted labels. 
We use a self-training approach~\citep{Jie2019} that iteratively estimates confidence scores in multiple training iterations and re-annotates the data at each iteration. 
Experiments show that our approach outperforms baselines on a general noisy-labeled setting with datasets in four languages and shows promising results on a distantly-labeled setting with four datasets.

\section{Method}
\label{sec:method}


Given a sentence $x = [x_1, ..., x_n]$ and its tag sequence $\hat{y}_1, ..., \hat{y}_n$, $n$ is the sentence length. 
We model the conditional probability of $y$ with a bi-directional LSTM-CRF~\citep{huang2015bidirectional}:
\begin{align}
    &h =\text{BiLSTM} (x) \quad \Phi_i = \text{Linear}(h_i) \\ 
    &p(y | x) = \Phi(y)/ Z \quad \alpha, Z= \text{Forward}(\Phi)
\end{align} 
Where $h$ denotes LSTM states, Linear$(\cdot)$ denotes a linear layer, $\Phi(y)$ denotes the potential (weight) evaluated for tag sequence $y$, $Z$ denotes the partition function, $\alpha$ denotes the forward variables, and Forward$(\cdot)$ denotes the Forward algorithm~\citep{sutton2006introduction}. 
The advantage of the CRF model is that it gives us a probabilistically uniform way to handle labels we do or do not trust by partial marginalization, which we discuss later. 
	
\subsection{Confidence Score Estimation}
\label{ssec:noise_detection}

Our confidence estimation model reuses the base LSTM-CRF architecture and
assigns a confidence score $s_i$ for each $\hat{y}_i$. 
A natural choice is to use the CRF marginal probability:
\begin{align}
    s_i = p(\hat{y}_i | x) \quad p(y_i | x) = \alpha_i \beta_i / Z
\end{align}
where $\beta$ is the backward variable and can be computed with the Backward algorithm~\citep{sutton2006introduction}.
This strategy infers $s_i$ based on global-normalization and assumes strong dependency between consecutive labels. 
The intuition is that annotators are more likely to make mistakes on a label if they have already made mistakes on previous labels. 

Our second strategy makes a stronger local independence assumption and considers a noisy label at step $i$ only relies on the word context, not the label context.  
To this end, we use a simple categorical distribution parameterized by a Softmax: 
\begin{align}
    s_i = p(\hat{y}_i | x) \quad p(y_i | x) = \text{Softmax}(\Phi_i)
\end{align}
Here we reuse the factor $\Phi_i$ as the logits of the Softmax because in the CRF context it also means how likely a label $y_i$ may be observed given the input $h_i$. 
Intuitively, this strategy assumes that annotators make mistakes solely based on words, no matter whether they have already made mistakes previously.

\begin{table*}
	\centering
	\small
	\begin{tabular}{@{}m{5.25cm}>{\centering}m{0.75cm}> {\centering}m{0.75cm}>{\centering}m{0.75cm}> {\centering}m{0.75cm}>{\centering}m{0.01cm}> {\centering}m{0.8cm}>{\centering}m{0.8cm}> {\centering}m{1cm}> {\centering\arraybackslash}m{1cm}}
		\hline\multirow{2}{*}{Method} & \multicolumn{4}{c}{General Noise} && \multicolumn{4}{c}{Distant Supervision}  \\ 
		\cline{2-5} \cline{7-10}
		 & En &  Sp &  Ge &  Du && CoNLL & Tweet  & Webpage &  Wikigold \\
		\hline
		 1. BiLSTM-CRF & 73.3 & 61.9 & 57.7 & 58.3 && 59.5 & 21.8 & 43.3 & 42.9 \\
		 2. BiLSTM-CRF (clean data upper bound) & 90.3 & 85.2 & 77.3 & 81.1 && 91.2 & 52.2  & 52.3 & 54.9\\
		 3. RoBERTa (clean data upper bound) & - & - & - & - && 90.1 & 52.2  & 72.4 & 86.4\\
		 \hline 
		 \multicolumn{8}{@{}l}{{\textit{Proposed for General Noise Setting}}} \\
		 4. NA \cite{hedderich-klakow-2018-training} & 61.5 & 57.3 & 46.1 & 41.5  && - & - & - & - \\
		 5. CBL \cite{mayhew2019named} & 82.6 & 76.1 & 65.6 & 68.5 && 75.4 & 18.2 & 31.7 & 42.6\\ 
		 6. Self-training (\citealp{Jie2019}) & \colorback{84.0} & \colorback{71.4} & \colorback{66.5} & \colorback{59.6} && \colorback{77.8} & \colorback{42.3} & \colorback{49.6} & \colorback{51.3} \\ \hline
		 \multicolumn{8}{@{}l}{{\textit{Proposed for Distant Supervision Setting}}} \\
		 7. AutoNER \cite{shang2018learning}  & - & - & - & -  && 67.0 & 26.1 & 51.4 & 47.5 \\ 
		 8. LRNT \cite{cao-etal-2019-low}  & - & - & - & - &&  69.7 & 23.8 & 47.7 & 46.2 \\ 
		 9. BOND (RoBERTa \citealp{liang2020bond}) & - & - & - & - && \bf 81.5 & \bf 48.0 & \bf 65.7 & \bf 60.1 \\ 
		 \hline
		 \multicolumn{8}{@{}l}{{\textit{Ours, best configurations}}} \\
		 10. Ours (local, $\tau^*$)  & \bf \colorback{87.0} & \colorback{78.8} &  \colorback{68.3} & \colorback{69.1} && \colorback{79.4} & \colorback{43.6} & \colorback{51.8} & \colorback{54.0} \\
		 11. Ours (global, $\tau^*$) & \colorback{86.4} & \colorback{79.0} & \bf \colorback{69.2} & \bf \colorback{71.2} && \colorback{79.2} & \colorback{43.1} & \colorback{50.0} & \colorback{53.0}  \\
		 \hline 
		 \multicolumn{8}{@{}l}{{\textit{Ours, other possible configurations}}} \\
		 12. Ours (local, $\tau^\star$)  & 86.2 & \bf 79.2 & 68.2 & 67.2 && - & - & - & - \\
		 13. Ours (global, $\tau^\star$) & 85.4 & 75.4 & 68.4 & 69.0 && -  & - & - & - \\
		 14. Ours (local, $\tau^*$, w/o. calibration) & \colorback{85.8} & \colorback{77.3} &  \colorback{67.2} & \colorback{68.0}  &&  \colorback{79.9} & \colorback{40.8} & \colorback{46.9} & \colorback{50.0} \\
		 \hline
		 \multicolumn{8}{@{}l}{{\textit{Ours with pretrained LM}}} \\
	     15. Ours (local, $\tau^*$, BERT) & - & - & - & - && 77.2& 46.7& 59.3 & 57.3 \\
		 16. Ours (global, $\tau^*$, BERT) & - & - & - & - && \colorback{78.9} & \colorback{47.3}  & \colorback{61.9} & \colorback{57.7}  \\

		\hline
	\end{tabular}
	\caption{Results (F1\%) on artificially perturbed datasets and distantly supervised datasets. $\tau^*$ = searched, $\tau^\star$ = oracle. }
	\label{tab:result}
\end{table*}

\subsection{Confidence Calibration and Partial Marginalization}
\label{ssec:partial_marginalization}

We use $s_i$ to decide if we want to trust a label $\hat{y}_i$ and marginalize out labels we do not trust. 
Our marginalization relies on a threshold 
to determine the portion of trusted labels and 
the noise ratio that we believe the data contain. 
Given a batch of $(x, \hat{y})$ pairs, after confidence estimation, we collect all word-label-confidence triples into a set $\mathcal{D} = \{x_j, \hat{y}_j, s_j\}_{j=1}^N$, $N$ denotes  total number of the triples. 
 
We further separate the estimation for positive labels (entities) and negative labels (i.e., the \textit{O} label) because we empirically observe that their probabilities are consistently different. 
To this end, 
we divide $\mathcal{D}$ into positive and negative groups $\mathcal{D}_\text{p} = \{(x_j, \hat{y}_j, s_j), \hat{y}_j \in \mathcal{Y}_\text{p}\}$ and $\mathcal{D}_\text{n} = \{(x_j, \hat{y}_j, s_j), \hat{y}_j \in \mathcal{Y}_\text{n}\}$, $\mathcal{Y}_\text{p}$ and  $\mathcal{Y}_\text{n}$ denotes sets of positive and negative labels. 
We rank triples in $\mathcal{D}_l$ ($l \in $ \{p, n\}) according to confidence scores and retain the most confident $r_l(e) \cdot |\mathcal{D}_l|$ triples at epoch $e$ as clean for which we do  maximum likelihood.
We view the remaining 
triples as noisy and marginalize them out. 
We update the keep ratio $r_{l}(e)$ at each epoch following~\citet{han-etal-2018-fewrel}: 
\begin{align}
r_l(e)=1-\min \left\{\frac{e}{K} \tau_l, \tau_l\right\}, l \in \{\text{p}, \text{n}\}
\end{align}
where $\tau_{l}$ is the ratio of noise that we believe in the training data.
Basically this says we gradually decrease the epoch-wise keep ratio $r_l(e)$ to the full ratio $1 - \tau_{l}$ after $K$ epochs. 
We grid-search $\tau_{l}$ heuristically in experiments (results in Figure~\ref{fig:subfig:b}).

\begin{figure}[t]
		\centering
		\includegraphics[width=3in]{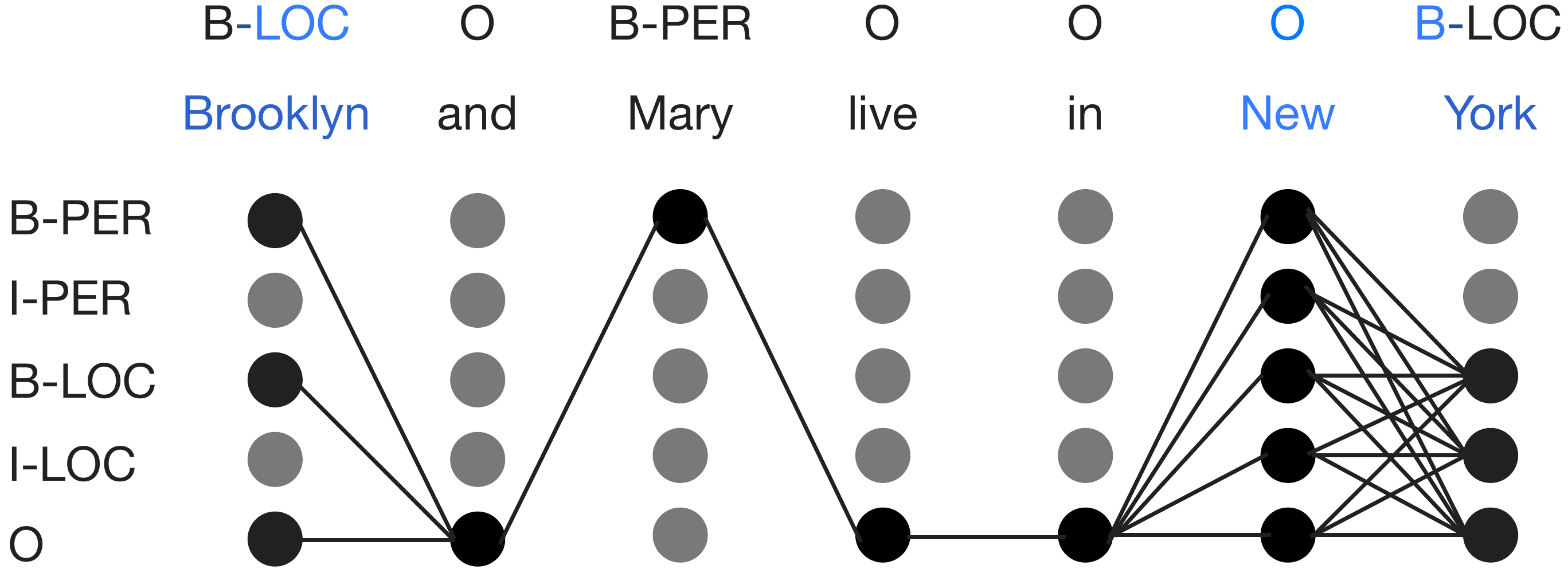}
		\caption{A partial marginalization example after confidence estimation.
		In this example, we do not trust any labels for \textit{New} (so we marginalize all labels out), partially trust labels for \textit{Brooklyn} (\textit{B} part) and \textit{York} (\textit{LOC} part, so we sum over labels we trust), and fully trust labels for the rest words (so we simply evaluate and maximize their weights.). 
		}
		\label{fig:crf}
	\end{figure}

For positive cases in $\mathcal{D}_\text{p}$ viewed as noisy according to the previous procedure, we do a further confidence calibration.
Noting that a $y_i$ always take the form $y_{i}^{{p}}$-$y_{i}^{{t}}$ (position-type) 
(e.g. if $y_{i}=\textit{B-LOC}$, then $y_{i}^{{p}}=\textit{B}$ and $y_{i}^{{t}}=\textit{LOC}$), 
an important assumption is that annotators are unlikely to mistake both parts --- mistakes usually happen on only one of them. 
So we calculate two calibrated confidence scores $s_i^p$ and $s_i^t$ for $\hat{y}_i^p$ and $\hat{y}_i^t$:
\begin{align}
s_i^p &=\frac{1}{|Y(\hat{y}_i^p)|} \sum_{y_i} p(y_i | x) \quad \text { where } y_{i}^p=\hat{y}_i^{p}  \\ 
s_i^t &=\frac{1}{|Y(\hat{y}_i^t)|}\sum_{y_i}  p(y_i | x) \quad \text { where } y_{i}^t=\hat{y}_i^{t}
\end{align}
where $Y(\hat{y}_i^t)$ denotes the set of labels sharing the same $\hat{y}_i^t$ part, and $Y(\hat{y}_i^p)$ is defined similarily.
If $s_i^p > s_i^t$, we trust the $\hat{y}_i^p$ (position) part of the label and marginalize out all labels with different positions except for the \textit{O} label. 
For example, in Figure~\ref{fig:crf}, for the word \textit{Brooklyn} we trust the all labels with the position \textit{B} (\textit{B-PER} and \textit{B-LOC}) and the \textit{O} label, sum over the tag sequences passing these labels, and reject other labels. 
Similar operation applies for cases where $s_i^p < s_i^t$ (E.g., the word \textit{York}).
For labels we do not trust in the negative group $\mathcal{D}_\text{n}$, we simply marginalize all labels out (E.g., the word \textit{New}). 
We maximize the partially marginalized probability~\citep{Bellare2007}: 
\begin{align}
    \tilde{p}(\hat{y} | x) = \sum_{y \in \tilde{Y}} \Phi(y) / Z \label{eq:partial_marginalization}
\end{align}
where $\tilde{Y}$ denotes the set of tag sequences compatible with $\hat{y}$ after confidence estimation. 
A concrete example is given in Figure~\ref{fig:crf}. 
The summation in equation~\ref{eq:partial_marginalization} can be calculated exactly with Forward-styled dynamic programming~\citep{Sasada2016}.

	
\subsection{Self Training}
\label{ssec:self_training}

We integrate our approach into a  self-training framework proposed by~\citet{Jie2019}. 
At each round, the training set is randomly divided into two parts for cross-validation.
We iteratively re-annotate half of the training set with a model trained on the other half.
After a round, we use the updated training set to train the next round.

\section{Experiments}
\label{sec:exp}

\begin{figure*}
\centering
\subfigure[Training curve]{\label{fig:subfig:a}\includegraphics[width=2.18in]{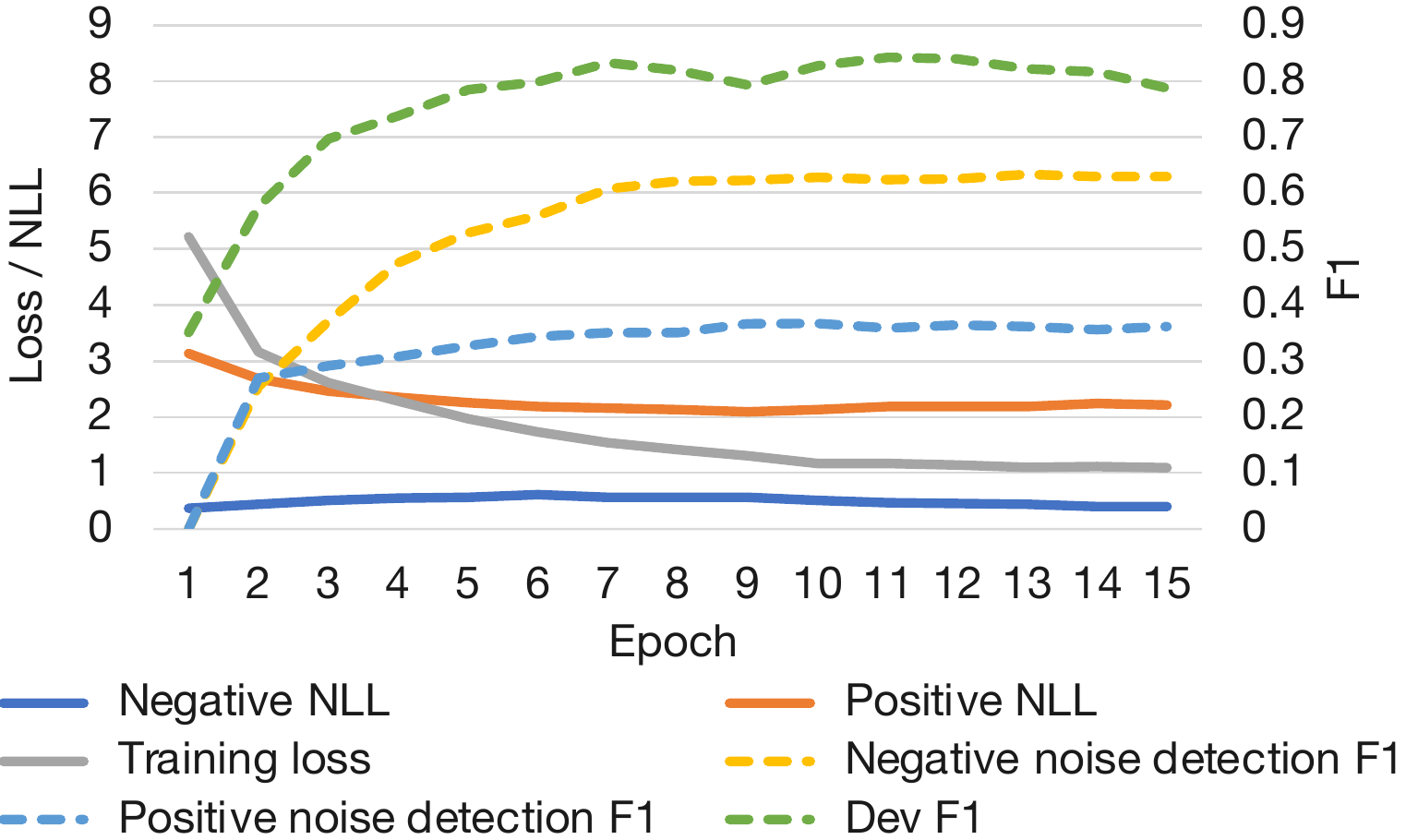}}
\subfigure[Noise rate search curve]{\label{fig:subfig:b}\includegraphics[width=2.2in]{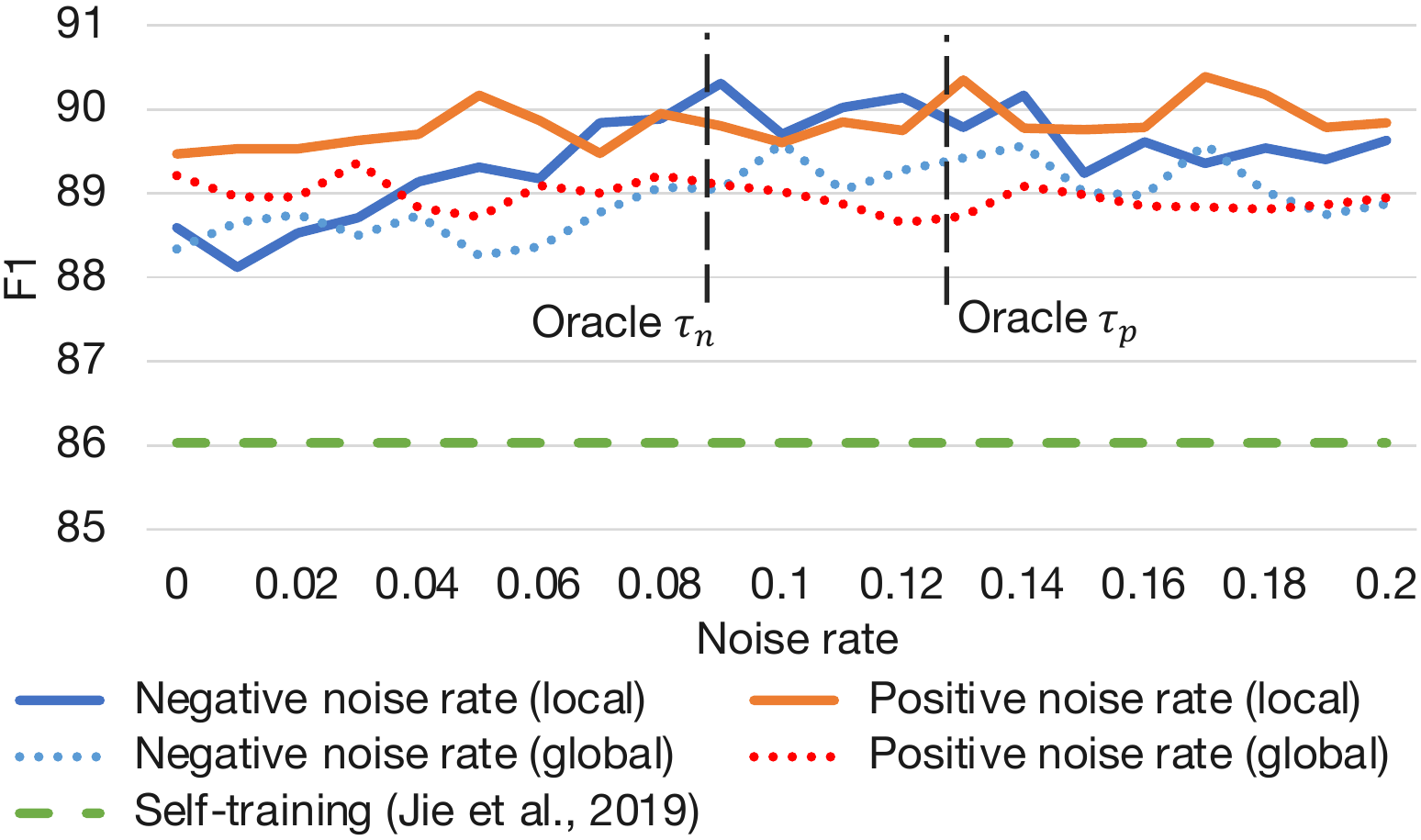}}
\subfigure[Results w.r.t. noise level ]{\label{fig:subfig:c}\includegraphics[width=1.7in]{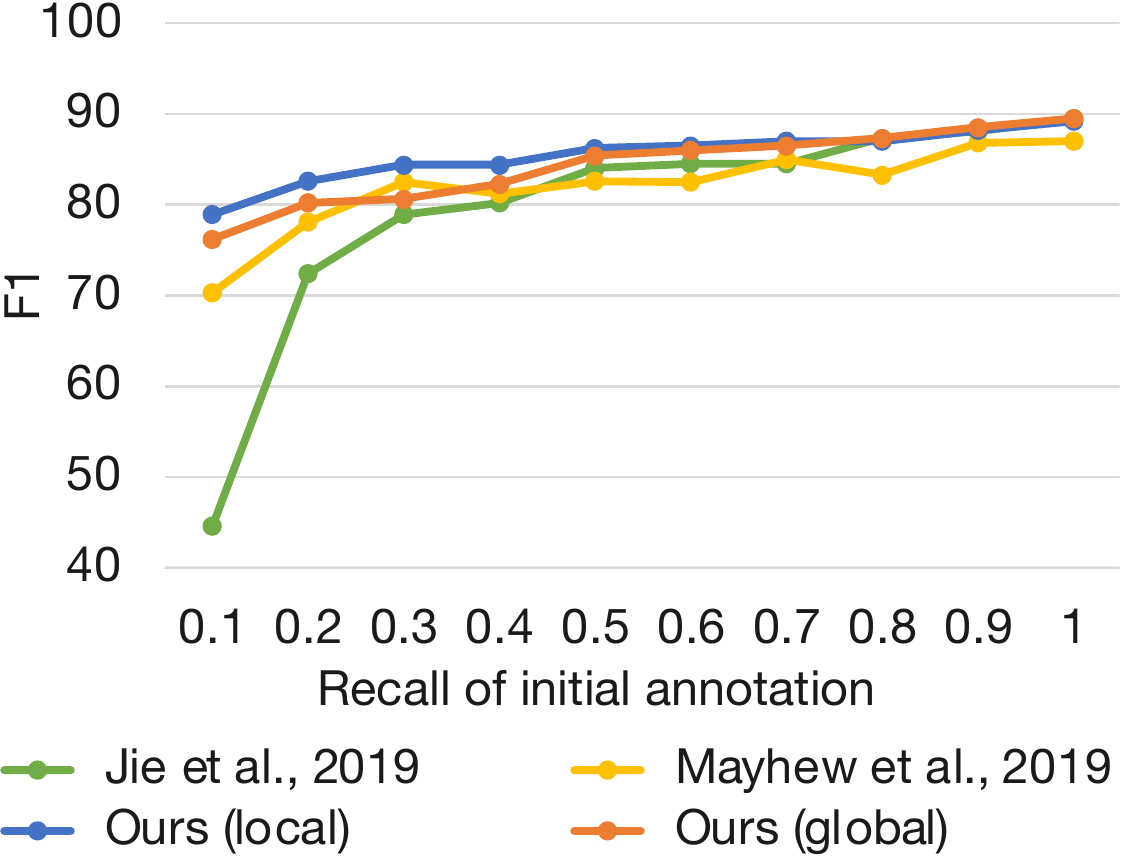}}
\caption{Analysis on English CoNLL03 dataset. (a) Dev performance strongly correlates to loss values (confidence scores) and noise detection performance. (b) An over-estimate of noise tends to give better performance. (c). Our approach is particularly effective under larger noise (lower recall = larger noise).}
\label{fig3}
\end{figure*}
\begin{figure}
\centering
 \includegraphics[width=2.8in]{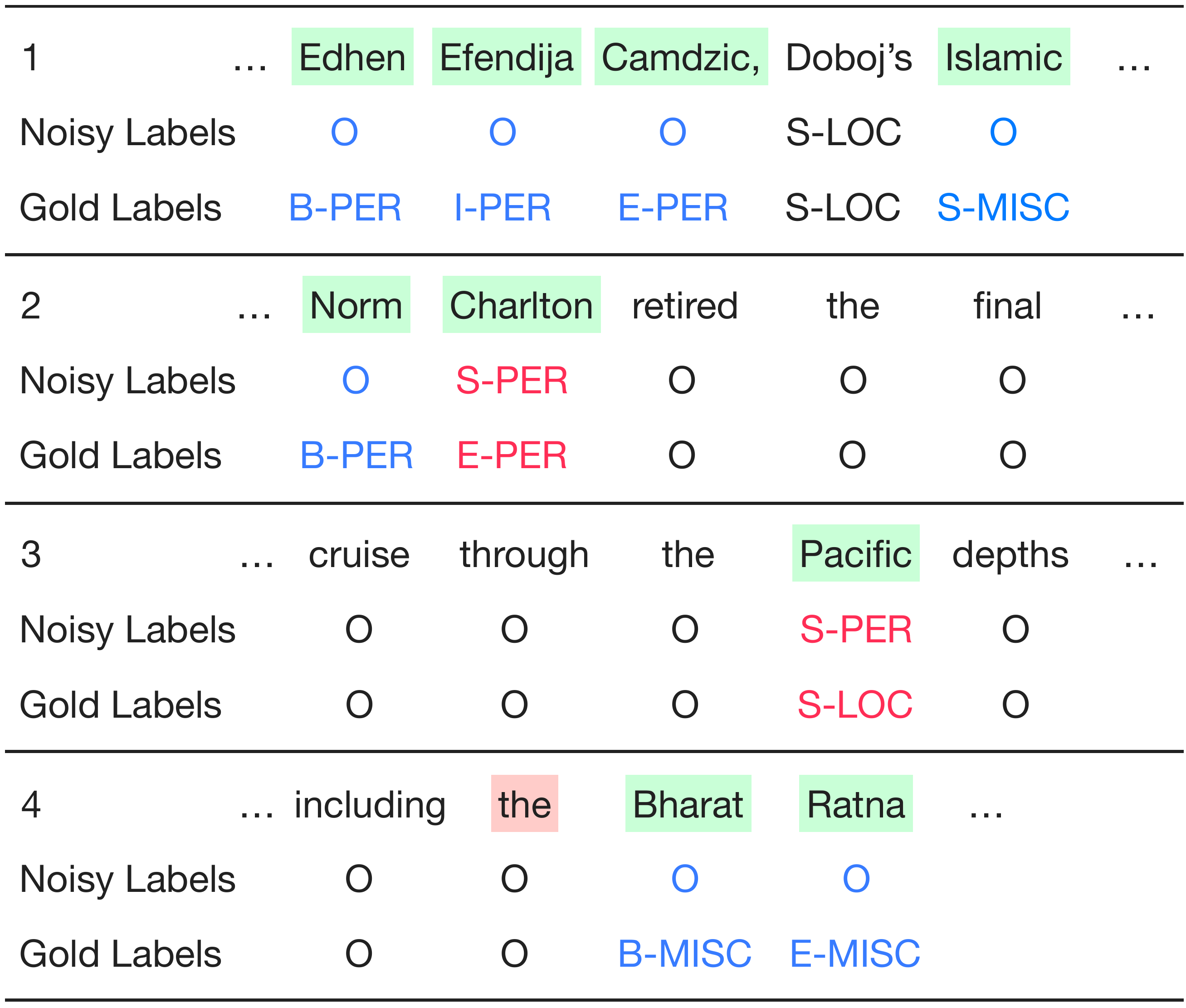}
  \caption{Confidence estimation case study. Red fonts = noisy positive, blue fonts = noisy negatives. Green shade = correct noise detection, red shade = wrong noise detection.
  }
 \label{case_study}
\end{figure}

\subsection{Datasets and Baselines}

\noindent \textbf{General Noise.} 
Following~\citet{mayhew2019named}, we first consider general noise by artificially perturbing the CoNLL dataset~\citep{sang2003introduction} on four languages including English, Spanish, German, and Dutch. 
Gold annotations are perturbed by: (a) tagging some entities to \textit{O} to lower the recall to 0.5; (b) introducing some random positive tags to lower the precision to 0.9. We compare our methods with Noise Adaption (NA, \citealp{hedderich-klakow-2018-training}), Self Training~\cite{Jie2019}, and CBL~\citep{mayhew2019named}. 
This setting is for testing our approach in a controlled environment. 

\noindent \textbf{Distant Supervision.}  We consider four datasets including CoNLL03~\citep{sang2003introduction}, Tweet~\citep{godin2015multimedia}, Webpage~\citep{ratinov2009design}, and Wikigold~\citep{balasuriya2009named}. 
In this setting, the distantly supervised tags are generated by the dictionary following BOND~\citep{liang2020bond}. 
We compare our methods with AutoNER~\citep{shang2018learning}, LRNT~\citep{cao-etal-2019-low}, and BOND. 
This setting aims to test our approach in a more realistic environment.

\subsection{Results}

Table~\ref{tab:result} shows our primary results. 
We use \textit{local} and \textit{global} to denote locally / globally normalized confidence estimation strategies. 
We use \textit{oracle} (unavailable in real settings) / \textit{searched} $\tau$  to denote how we obtain the prior noise ratio $\tau$. 
We note that the Self-training baseline~(\citealp{Jie2019}, line 6) is the most comparable baseline since our confidence estimation is directly integrated into it. 
We primarily compare this baseline with our best configurations~(line 10 and 11). 
We focus on the\colorback{shaded} results as they are the most informative for demonstrating our method. 

\noindent \textbf{General Noise.} Our methods (both local and global) outperforms the state-of-the-art method~\citep{Jie2019} by a large margin in three datasets (En, Sp, Du, line 10 and 11 v.s. 6), showing the effectiveness of our approach. 
We observe the oracle $\tau$ does not necessarily give the best performance and an over estimate of confidence could leave a better performance. 
Ablation results without calibration further show the effectiveness of our calibration methods (line 10 v.s. 14). 
We note that the CoNLL dataset is an exception where the calibration slightly hurts performance. 
Otherwise the improvements with calibration is clear in the other 7 datasets.

\noindent \textbf{Distant Supervision.} 
Our method outperforms AutoNER and LRNT without pre-trained language models.
Reasons that we are worse than BOND (line 16 v.s. 6) are: (a) many implementation aspects are different, and it is (currently) challenging to transplant their settings to ours; (b) they use multiple tailored techniques for distantly-labeled data (e.g., the adversarial training), while ours is more general-purpose. 
Though our method does not outperform BOND, it still outperforms AutoNER and LRNT (under the setting all without pretrained model, line 10 and 11 v.s. 7 and 8) and shows promising gain.

\subsection{Further Analysis}
We conduct more detailed experiments on the general noise setting for more in-depth understanding.

\noindent \textbf{Training Dynamics (Figure~\ref{fig:subfig:a}).} 
As the model converges, as clean data converge faster, the confidence gap between the clean and the noisy is larger, thus the two are more confidently separated, so both noise detection F1 and dev F1 increase.

\noindent \textbf{Noise Rate Search (Figure~\ref{fig:subfig:b}).}
Our method consistently outperforms baseline without confidence estimation. 
Lines tend to be higher at the right side of the figure, showing
 an over-estimate of noise tends to give better performance. 

\noindent \textbf{Level of Noise (Figure~\ref{fig:subfig:c}).} 
In many real-world scenarios, the noise w.r.t. precision is more constant and it is the recall that varies. 
So we simulate the level of noise with different recall (lower recall = larger noise ratio).
Our method outperforms baselines in all ratios and is particularly effective under a large noise ratio. 

\noindent \textbf{Case Studies (Figure \ref{case_study}).}
The top three cases give examples of how our method detects: (1)~false negative noise when an entity is not annotated, (2)~entities with wrong boundaries and (3)~wrong entity types. 
The last example (case 4) gives a failure case when the model treats some correct tags as noise due to our over-estimate of noise (for better end performance). 

\section{Related Works}
\label{sec:related_works}
State-of-the-art NER models \cite{ma2016end,Lample2016,Devlin2018} are all under the traditional assumption of clean data annotation. 
The key motivation of this work is the intrinsic gap between the clean data assumption and noisy real-world scenarios. 
We believe that the noisy label setting is a fundamentally challenging in NER and all related supervised learning tasks.

Previous works on NER with noise could be organized into two threads: (a) some works treat this task as learning with missing labels. \citet{Bellare2007} propose a missing label CRF to deal with partial annotation. \citet{Jie2019} propose a self-training framework with marginal CRF to re-annotate the missing labels. (b) other works treat missing labels as noise and try to avoid them in the training process. For example,
\citet{mayhew2019named} train a binary classifier supervised by entity ratio to classify tokens into entities and non-entities.

A widely-used way to collect NER annotations is distant supervision, which consequently becomes an important source of noise.
\citet{peng2019distantly} formulate this task as the positive-unlabeled (PU) learning to avoid using noisy negatives. AutoNER \cite{shang2018learning} trains the model by assigning ambiguous tokens with all possible labels and then maximizing the overall likelihood using a fuzzy LSTM-CRF model.~\citet{cao2019low} and~\citet{Yang2018} try to select high-quality sentences with less annotation errors for sequential model.~\citet{liang2020bond} leverage pre-trained language models to improve the prediction performance of NER models under a self-training framework.

Our inspiration of confidence estimation comes from the so-called 
memorization effect observed in the computer vision \cite{jiang2018mentornet,han2018coteaching,arazo2019unsupervised}.
It observes that neural networks usually take precedence over noisy data to fit clean data, which indicates that noisy data are more likely to have larger loss values in the early training epochs \cite{10.5555/3305381.3305406}. In this work, we leverage it to estimate the confidence scores of labels.


\section{Conclusion}
\label{sec:conclustion}
In this work, we propose a calibrated confidence estimation approach for noisy-labeled NER. 
We integrate our method in an LSTM-CRF model under a self-training framework. 
Extensive experiments  demonstrate the effectiveness of our approach. 
Our method outperforms strong baseline models in a general noise setting (especially for larger noise ratios), and shows promising results in a distant supervision setting.

\section{Acknowledgments}
We thank all anonymous reviewers for their helpful comments. This work is supported by Alibaba Group through Alibaba Research Intern Program and AZFT Joint Lab for Knowledge Engine.

\bibliographystyle{acl_natbib}
\bibliography{anthology}

\clearpage
\appendix
\begin{table*}[t!]
	\centering
	\begin{tabular}{l|cc|cc|cc}
		\hline
		\multirow{2}{*}{ Dataset} & \multicolumn{2}{c|}{ Training} & \multicolumn{2}{c|}{ Dev} & \multicolumn{2}{c}{ Test} \\
		& \#entity & \#sent & \#entity & \#sent & \#entity & \#sent\\
		\hline
		 English & 23,499 & 14,041 & 5,942 & 3,250 & 5,648 & 3,453  \\
		 Spanish& 18,796 & 8,322 & 4,338 & 1,914 & 3,559 & 1,516 \\
		 German & 11,851 & 12,152 & 4,833 & 2,867 & 3,673 & 3,005  \\
		 Dutch & 13,344 & 15,806 & 2,616 & 2,895 & 3,941 & 5,195  \\
		 CoNLL & - & 14,041 & - & 3,250 & - & 3,453   \\
		 Tweet & - & 2,393 & - & 999 & - & 3,844   \\
		 Webpage & - & 385 & - & 99 & - & 135  \\
		 Wikigold & - & 1,142 & - & 280 & - & 274  \\
		\hline
	\end{tabular}
	\caption{Statistics of datasets.}
	\label{tab:dataset}
\end{table*}

\section{Dataset Processing}

\subsection{Artificially Perturbed Dataset}
The gold annotations of training data are perturbed by lowering the recall and precision following \citet{mayhew2019named}. Firstly, we randomly select an entity from the whole entity set and tag all of its occurrences to `O'. We repeat this operation until the recall decreases to 0.5. Then, we randomly tag some tokens/spans to the entity label to decrease the precision to 0.9. The detailed data statistics are shown in Table \ref{tab:dataset}. 

\subsection{Distarntly Supervised Dataset}
All distantly supervised datasets in our experiments are the same as those in \citet{liang2020bond}. The distant labels are generated by external knowledge bases (e.g. Wikidata \citealp{vrandevcic2014wikidata}) and gazetteers collected from multiple online resources. Specifically, the entity candidates are first detected by POS tagger (NLTK \citealp{loper2002nltk}). Next, the ambiguous candidates are filtered out by the Wikidata query service. Then, they match the entities with words in multi-resources gazetteers to get their entity types. Additional rules are used to get the entity labels of the unmatched tokens. The detailed data statistics are shown in Table \ref{tab:dataset}. 

\section{Implementation Details}

\subsection{Model Structure and Implementation}

For all the experiments with LSTM, we use the same word embeddings as \citet{Lample2016}. We use the character-level LSTM with hidden size 25 to produce character-level word embeddings. The concatenation of the two embeddings are fed into BiLSTM with hidden size 100. 
We also apply the dropout \cite{Srivastava2014} between layers, with a rate of 0.5. The model is optimized using Stochastic Gradient Descent \cite{robbins1951stochastic} with a learning rate of 0.01. 

For experiments with BERT, we use the BERT-base \cite{Devlin2018} as our encoder. The implementation is based on the codebase HuggingFace Transformers \cite{wolf-etal-2020-transformers}. The dropout is set to 0.2. The model is optimized using Adam \cite{Kingma2014} with an initial learning rate of 3e-5.

\subsection{Hyper-Parameters}

There are two important hyper-parameters in our model as the positive noise rate $\tau_{p}$ and the negative noise rate $\tau_{n}$. Based on our observation, the initial noise rates are various in different datasets. However, since our model has the ability to handle the noise, the noise rates are relatively stable after the first iteration of self training. Therefore, we empirically set $\tau_{p}$ and $\tau_{n}$ to 0.005 and 0.15 for all experiments from the second iteration. For the first iteration, we report the results of two strategies as follows:

\noindent \textbf{Oracle.} 
`Oracle' means that we use the gold noise ratio (unavailable in real settings) of positive noise rate $\tau_{p}$ and negative noise rate $\tau_{n}$. The strategy is only applicable for artificially perturbed datasets since the completed annotation is known.

\noindent \textbf{Searched.} `Searched' means that we search the two hyper-parameters for best performance on the development set. We search two parameters separately since we assume $\tau_{n}$ and $\tau_{p}$ are independent. The search ranges from 0.0 to 0.2 with an interval of 0.01. We determine the two parameters with the best development result on different datasets.

\section{Analysis of Self Training}
\begin{figure}
 \includegraphics[width=3in]{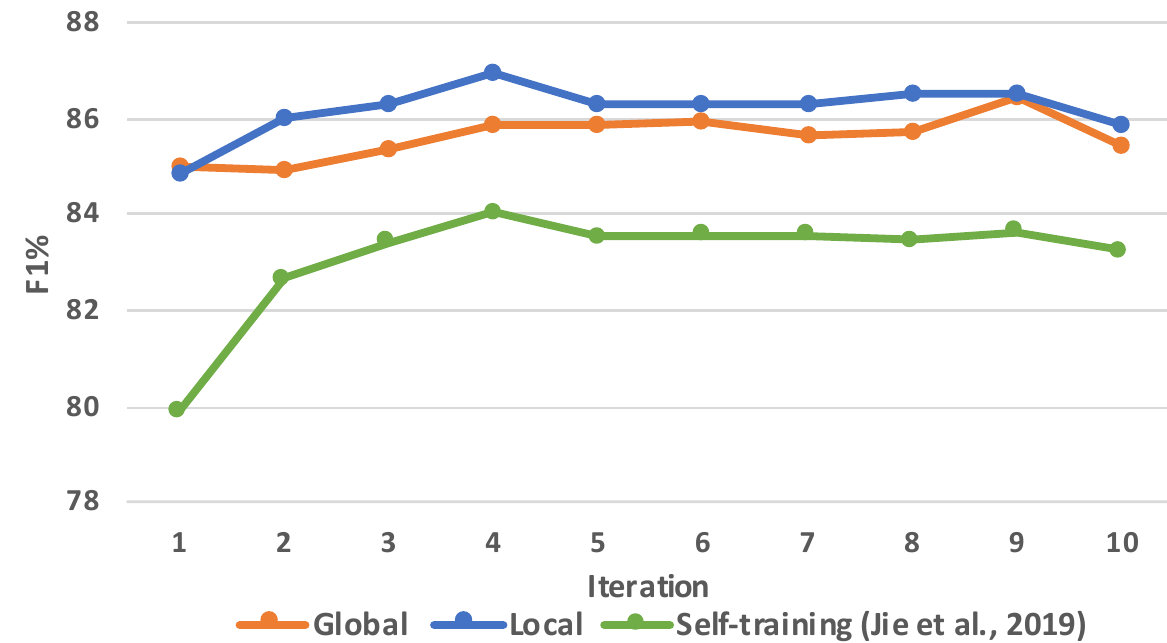}
  \caption{Results of self-training.}
 \label{self-training}
\end{figure}

The self training is borrowed from \citet{Jie2019} and not our main contribution. However, to be self-contained, we also report the results of self training in Figure~\ref{self-training}. Our method (both local and global) outperforms the baseline by a large margin at the first iteration, which indicates we have a better base model of handling noise. Also, all curves raise in the first several iterations and maintain stable relatively in the subsequent iterations. 

\end{document}